\def\year{2021}\relax
\documentclass[letterpaper]{article} 
\usepackage{aaai21}  
\usepackage{times}  
\usepackage{helvet} 
\usepackage{courier}  
\usepackage[hyphens]{url}  
\usepackage{graphicx} 
\usepackage{graphicx}  
\usepackage{natbib}  
\usepackage{caption} 
\usepackage{comment}
\usepackage{mathtools}
\usepackage{makecell}
\usepackage{amssymb}
\usepackage{amsmath}
\usepackage{kotex}
\usepackage{multirow,hhline,array}
\usepackage[switch]{lineno}  %

\usepackage[usenames, dvipnames]{xcolor}

\title{On the Orthogonality of Knowledge Distillation with Other Techniques: \\
From an Ensemble Perspective}

\author{
    SeongUk Park, KiYoon Yoo, Nojun Kwak\\
}
\affiliations{
    \textsuperscript{\rm}Graduate School of Convergence Science and Technology\\
      Seoul National University\\
      \texttt{\{swpark0703, 961230, nojunk\} @snu.ac.kr}\\

}

\begin{document}

\maketitle

\begin{abstract}
To put a state-of-the-art neural network to practical use, it is necessary to design a model that has a good trade-off between the resource consumption and performance on the test set. 
Many researchers and engineers are developing methods that enable training or designing a model more efficiently. 
Developing an efficient model includes several strategies such as network architecture search (NAS), pruning, quantization, knowledge distillation, utilizing cheap convolution, regularization, and also includes any craft  that leads to a better performance-resource trade-off. 
When combining these technologies together, it would be ideal if one source of performance improvement does not conflict with others.
We call this property as the \textit{``orthogonality"} in model efficiency. In this paper, we {focus on knowledge distillation and demonstrate that knowledge distillation methods are \textit{orthogonal} to} other efficiency-enhancing methods both analytically and empirically. 
Analytically, we claim that knowledge distillation function{s} analogous to a model ensemble method, bootstrap aggregating {(bagging)}. 
This analytical explanation is provided {from the perspective of (implicit) data augmentation property of} knowledge distillation.
Empirically, we verify knowledge distillation as a powerful apparatus for practical deployment of efficient neural network, and also introduce ways to integrate it with other methods effectively.
\end{abstract}

\section{Introduction}
Developing an efficient deep neural network (DNN) is an important issue {enabling} DNNs to be utilized in {cutting-edge} devices. 
Many researchers {have proposed} methods in {making efficient} DNNs from various perspectives, and achieved meaningful results. 
These encompass (but are not restricted to) the following representative fields: 1) knowledge distillation \cite{hinton2015distilling} which guide{s a} small-capacity student network using outputs from one or more high-capacity teacher networks, 2) network pruning \cite{han2015deep,anwar2017structured,li2016pruning} which erases disposable weights in an over-parameterized network, 
3) building an efficient model such as MobileNet \cite{howard2017mobilenets}, SqueezeNet \cite{iandola2016squeezenet}, NAS \cite{zoph2016neural}, EfficientNet \cite{tan2019efficientnet} which attempts to make parameter-efficient neural network cells or blocks, and 4) weight quantization \cite{rastegari2016xnor,gong2014compressing,courbariaux2015binaryconnect} which restricts the bit-width of the parameters. 
In addition, 5) regularization techniques also play a role of enhancing parameter-efficiency 
in that the model benefits on the test accuracy without having to use a larger network.

{Given the array of methods, it begs the question: 
\textit{If each of them operates by a distinct mechanism, can we expect better results by using them simultaneously?}} 
It would be ideal if these methods can be combined into a single model without degrading the performance or efficiency gain of each.
{Throughout this paper,} we {denote} this property as the {``\textbf{orthogonality}" in} model efficiency. 
Discovering the orthogonal property between the methods would lead to a practicable solution of deploying deep learning algorithms into real-world problems.
Among various approaches, in this paper, we focus on knowledge distillation (KD) and seek ways to adapt it in parallel with other techniques by focusing on the characteristics of KD that functions analogous to \textit{model ensemble}{, especially} \textit{bagging}. Because model ensemble on average attains performance improvement when combined with other methods {\cite{Goodfellow-et-al-2016}}, we hypothesize that KD will inherit this {orthogonality to} other efficiency-improving methods. 
We empirically verify this by applying KD in combination with network pruning methods and regularization techniques. 

The rest of the paper is organized as follows.
{Some closely related research topics} are described in  Section \ref{sec:rel}. 
{Section \ref{sec:kd}} explains our analysis of how KD resembles bagging.
In Section \ref{sec:4} and \ref{sec:5},
we empirically demonstrate the orthogonal property of knowledge distillation between commonly used efficiency-improving methods: 1) network pruning, which is categorized as network compression methods and 2) several kinds of the image-level regularization (augmentation) methods. 
In Section \ref{sec:6}, we present qualitative experiments that provide some potential hints about the underlying mechanism of KD.
In the discussion section, we mention more about other efficiency-improving methods that we did not empirically handle in this paper. 
The detailed experiment settings for reproduction are included in the {supplementary materials.}

\section{Related Works}
\label{sec:rel}
In this paper, we report results {of} combining knowledge distillation with other model-efficiency improving methods.
Thus, in this section, we explain the preliminary studies that are closely related to our research topic. 

\subsection{Knowledge Distillation (KD)}
 
Throughout this paper, to avoid the possible controversy coming from the choice of different network types, we assume that KD is done between networks of the same type, instead of using different student-teacher network pairs
\footnote{
Empirically, stronger teacher is not always beneficial \cite{zhang2017deep}, and distillation to {the} same kind has been proven to perform fairly well in Born Again Neural Network (BAN) \cite{furlanello2018born} and FEED \cite{parkfeature}.
}. 
We divide KD into two categories: feature distillation and label distillation. 
We take both label distillation and feature distillation into consideration to observe whether feature distillation and label distillation behave differently when combined with other efficiency-improving methods.

\textbf{Label-based KD \ } 
During the nascent stage of KD research, \citet{ba2014deep} proposed to use MSE loss between the predicted labels of large and small models.
{Later on}, \citet{hinton2015distilling} proposed to learn to match the distributions of classification labels using KL-divergence instead of MSE, which is still used as a typical objective function for label-based KD.
In this paper, we will use the abbreviation VKD (vanilla KD) to refer to the method of Hinton's KL-Divergence that uses softened softmax logits.
The loss function of VKD is defined as  
\begin{equation}
\small
\mathcal{L}_{KD}=(1-\alpha)\mathcal{L}_{CE}(y, \sigma(\textbf{s})) + \alpha T^{2}\mathcal{L}_{KL}\bigg(\sigma\big(\frac{\textbf{s}}{T}\big), \sigma\big(\frac{\textbf{t}}{T}\big)\bigg),
\label{eq:LKD}
\end{equation}
where $\sigma(\cdot)$ refers to the softmax function, $\textbf{s}$ and $\textbf{t}$ each refers to the final logit of the student network and the teacher network. 
We use this method as a baseline since it still shows strong performance. 
\textbf{Feature-map based KD \ } 
Inspired by VKD, some studies have endeavored delivering useful information at the feature map level instead of distilling at the level of class label. 
Attention transfer (AT) \cite{zagoruyko2016paying} defined the channel-sum of activation of the feature map as an attention map, and achieved a performance improvement similar to that of VKD by matching the attention map of the teacher and the student. 
FitNet \cite{romero2014fitnets} is the first paper to propose the idea of a \textit{regressor} which is still utilized in most feature distillation methods. The regressor is a 1$\times$1 convolution layer made to cope with the discrepancy between the sizes of the feature maps caused by using different student and teacher networks.
Though many feature distillation loss terms have been proposed \cite{kim2018paraphrasing,heo2019comprehensive}, most of them stem from the regressor, which is even used beyond classification tasks \cite{chen2017learning,wang2019distilling}. 
In our paper, we use the following simple form of feature distillation loss:
\begin{equation}
\mathcal{L}_{FD}=\mathcal{L}_{CE}(y, \sigma(\textbf{s})) + \beta\|\frac{\phi_T}{\|\phi_T\|_2}-\frac{R(\phi_S)}{\|R(\phi_S)\|_2} \|_1,
\label{eq:LFD}
\end{equation}
where $\phi_T$ and $\phi_S$ each refers to the last feature map of the teacher network and the student network, and the $R(\cdot)$ is the regressor network. 
The hyperparameter $\beta$ balances the scale of the two losses\footnote{The choice of $\beta$ is explained  in the supplementary materials.}. 

\subsection{Network Pruning}
In this paper, we apply KD in combination with neural network pruning, which can be categorized into \textit{unstructured pruning} and \textit{structured pruning} depending on the pruned units. 

\textbf{Unstructured weight pruning \ } 
Unstructured pruning treats weights individually, and prunes out unimportant units at the weight-level.
Early studies such as Optimal Brain Damage \cite{lecun1990optimal} and Optimal Brain Surgeon \cite{hassibi1994optimal} proposed to eliminate redundant weights in a fully-connected network using second-order information. \citet{han2015learning} proposed to prune weights with small magnitude, and applied pruning to CNNs, such as AlexNets \cite{krizhevsky2012imagenet} and VGGs \cite{simonyan2014very}. 
Unstructured pruning is an effective way to reduce the model size as it can eliminate a large percentage of the parameters. 
The main drawback, however, is that although the number of parameters drastically decreases, the weights are contained in sparse matrices which do not contribute to actual inference speedup without a specialized hardware. 

\textbf{Structured pruning \ } 
Unlike unstructured pruning, structured pruning removes parameters by larger units such as channels (filters), layers, or blocks. 
\citet{li2016pruning} removed certain percentage of the convolution filters with smaller $L_1$ norm value at each layer.
Network Slimming \cite{liu2017learning} applies $L_1$ norm penalty at the channel-level during training of CNNs. 
The $L_1$ norm penalty induces sparsity at the channel-level enabling efficient pruning. 
At the pruning phase, the filters are sorted and eliminated by a certain global threshold.
\citet{huang2018data} applied this scheme to a higher level, such as residual blocks \cite{he2016deep} or groups of convolution layers.

\subsection{Regularization}

The main purpose of regularization is to enhance the generalization ability of a neural network. Any strategy that attempts to raise the test accuracy can be interpreted as regularization.
Here we introduce recent or commonly-used regularization methods.

\textbf{Weight-level regularization \ } 
Imposing {$L_2/L_1$} norm penalty {on} the weight parameters is a widespread practice: 
\begin{equation}
{J}{(\pmb{\theta})}=\mathcal{L}_{train} + \lambda ||\pmb{\theta}||_p^p, \quad p = 1 \text{ or } 2. 
\end{equation}
This term prevents the weights $\pmb{\theta}$ from becoming too large, restricting the representation power of the model.

\textbf{Image-level regularization \ } 
Various image data augmentation techniques have been widely used in practice: pixel normalization, horizontal flip, rotation, crop with paddings, pixel jittering, etc. 
AutoAugment \cite{cubuk2018autoaugment} proposes to {learn} the hyperparameters for {augmentation} strategies using reinforcement learning \cite{bello2017neural,zoph2016neural}, and Fast Auto-augment \cite{lim2019fast} boosts the hyperparameter {learning} process.
Cutout \cite{devries2017improved} proposes to mask some regions of the {training dataset} to {enhance} the robustness of a neural network. 
This {method} can be used in conjunction with {other} forms of data augmentations such as AutoAugment.

Mixup \cite{zhang2017mixup} linearly interpolate{s} both the {inputs} and the targets of two different training images, and improves the accuracy on state-of-the-art networks. 
Following these studies, Cutmix \cite{yun2019cutmix} proposes to augment train{ing} samples in {a} patch-wise manner. 
The image patches are cut and swapped between two training samples ($x_A$ and $x_B$), and simultaneously the ground truth labels ($y_A$ and $y_B$) are also mixed proportionally, creating an intermixed sample with its label being:
\begin{equation}
\tilde{x} = M \odot x_A + (1-M) \odot x_B, 
\label{eq2}
\end{equation}
\begin{equation}
\tilde{y} = \lambda y_A + (1-\lambda) y_B,  
\label{eq3}
\end{equation}
where {$\odot$ denotes pixel-wise multiplication}, $M \in \{0, 1\}^{W\times H}$ {is a mask} and the {mixing parameter} $\lambda$ is sampled from beta distribution $Beta(1,1)$. 

While Mixup, Cutout, and Cutmix all improve classification accuracy, only {the} CutMix-trained backbone model {has been} reported to improve detection and segmentation scores.
In the experiments of our paper, we {choose} AutoAugment and Cutmix to analyze the orthogonality of regularization methods with knowledge distillation.


\section{Orthogonality of KD in Model Efficiency}
\label{sec:kd}

\subsection{Orthogonal Property of Model Ensemble}

In model ensemble, if the errors of the individual models are perfectly correlated, model averaging {would essentially have} no effect. However, in the ideal case where the errors are perfectly uncorrelated, the expected squared error of the ensemble decreases linearly with the ensemble size
\cite{Goodfellow-et-al-2016}. 
{For} a neural network, it is a generally{-}held belief that the separately{-}trained networks will not make the same error due to random initialization, stochastic sampling, etc, even for the same kind of networks. This means on average, the ensemble will perform at least as high as one of its member. 
Though counterexamples {do exist in rare cases}, in practice model ensemble is a strategy often used to conveniently attain improvement in performance regardless of the underlying method. We call this characteristic as the orthogonal property of model ensemble.

\subsection{Ensemble Property of Knowledge Distillation}
VKD \cite{hinton2015distilling} initially proposed to use an ensemble model as the teacher so as to make the student network learn the ensemble knowledge. 
Though VKD proposed using an ensemble teacher, we infer that the use of KL-Divergence term in the KD loss function {itself} attributes a property of model ensemble even with a single-model teacher.
The term `property' is used to emphasize {that the two methods are not indeed identical}. 
Notably, model ensemble requires individual training and inference of every member, which does not {contribute} to model efficiency.

The ground of our analysis stems from the viewpoint of each training sample.
For a $C$-class classification problem, a single sample is allocated a one-hot label {$\textbf{t} \in \{0,1\}^C$} as the target. 
The cross-entropy (CE) loss for a single sample with an output logit $\textbf{q} \in \mathbb{R}_{+}^{C}$ is:
\begin{equation}
l_{CE} = -\sum^{C}_{i=1}{t}_{i}\log({q}_{i}) = -\log(q_{j}),
\label{eq:CE}
\end{equation}
%
where {the} groundtruth class {index for that sample is $j$}. 
In this case, the error function has a value only at the $j^{th}$ class as shown in the right term in (\ref{eq:CE}).
However, in VKD, where KL-divergence between the logit of the teacher ($\mathbf{p}$) and that of the student ($\mathbf{q}$) is used as the loss term, all classes contribute:
\begin{equation}
D_{KL}(\mathbf{p}|\mathbf{q}) = -\sum^{C}_{i=1}{p}_{i}\log({q}_{i}) + \sum^{C}_{i=1}{p}_{i}\log({p}_{i}).
\label{eq:KL}
\end{equation}
Considering KD of training the student with a fixed teacher network, the entropy term of (\ref{eq:KL}) is fixed and only the first cross entropy term has effect on the training of the student network.
This lends a new interpretation of KD in the perspective of data augmentation and model ensemble.
The original CE term in (\ref{eq:CE}) utilizes the original human-annotated training data $D_H = \{(x_{n},\textbf{t}_{n}) _{n=1}^N \}$. On the other hand, the CE term in (\ref{eq:KL}) can be interpreted as utilizing teacher-annotated training data $D_T = \{(x_{n}, \textbf{p}_{n})_{n=1}^N \}$. 
Note that $\textbf{p}_{n} \in [0,1]^C$ is a multi-class target vector. 
If we stick to the notion of one-hot target vector, $D_{T}$ can be interpreted as a new augmented dataset $D_{T}' = \{(x_{n}, (\textbf{c}_{i,n}, p_{i,n}) _{i=1}^C)  _{n=1}^{N} \}$ where $\textbf{c}_{i}$ is a one-hot vector with $i^{th}$ element being 1 and $(x, \textbf{c}, p)$ denotes a sample $(x, \textbf{c})$ with a fractional count $p$. 
Consequently, KD increases the effective number of samples.

Assuming that there exist $D_A \supset ( D_H \cup D_T )$, a dataset containing all the original and possible augmented dataset,
the training process of KD can be interpreted as {using subsets $D_{H}$ and $D_{T}$ from the set $D_{A}$}. 
Suppose two separate networks are trained, each using the {subset} $D_{H}$ and $D_{T}$ as the training dataset. 
If the output of the two models are averaged, it can be considered as bootstrap aggregating.
Instead of taking an average of two independently trained models' outputs, we let a single student network learn the weighted average of both targets to mimic the effect of bagging by using the KD loss term. 
Remember that we previously stated that model ensemble has the property of orthogonality.
Following this reasoning given any underlying efficiency-improving method, we expect applying KD\footnote{In this part, we only considered label distillation. However, for feature distillation, since there exists only linear operation between the logit and the last feature map, we expect it to behave similarly to label distillation.} would enhance the performance without conflicting with the performance gain of the method.
This view also explains many interesting phenomenon such as performance gain when using student and teacher networks of the same kind, which will be discussed further on Section \ref{sec:6}.

\begin{table*}[ht]
\begin{center}
\resizebox{1.0\linewidth}{!}{
\begin{tabular}{c|c|c|c|cccc|c}
\Xhline{3\arrayrulewidth}
Model Type                & Remain{ing} Channels         & Train Type & Unpruned  & Finetuned / Pre-Distill  & Post-Distill & Pre-Post & Self-Distill & Surplus  \\
\cline{1-9}
 \multirow{6}*{ResNet-56} & \multirow{3}*{40\% / 70\% / 90\%}& 
               Scratch    & 71.23   & 70.06  &   -   &   -   &   -   &   -  \\
        & &    Label      & 73.76(+2.53)   & 72.05  & 71.37 & 73.24  & \textbf{73.30}(+3.24) & 0.71   \\
        & &    Feature    & 73.60(+2.37)   & 71.27  & 70.93 & 72.44  & \textbf{72.81}(+2.75) &   0.38   \\
\cline{2-8}
                          & \multirow{3}*{50\% / 60\% / 70\%}& Scratch    & 71.23      & 68.55                   &   -          &   -    &-   &     \\
                          &                       & Label      & 73.76   & 70.58                   & 70.49        & 72.51  & \textbf{72.58}(+4.03)   &  1.50 \\
                          &                       & Feature    & 73.60   & 69.97                   & 69.83        & 71.76  & \textbf{71.81}(+3.26)  &   0.89 \\
\cline{1-8}
 \multirow{6}*{ResNet-110}& \multirow{3}*{80\% / 80\% / 80\%}& Scratch    & 72.76    & 71.71                   & - & -  &   -        \\
                          &                       & Label      & 74.82(+2.06)    & 72.68                   & 72.28        & 74.25  & \textbf{74.40}(+2.69)  &   0.63 \\
                          &                       & Feature    & 74.98(+2.22)   & 73.06                   & 72.41        & 74.15  & \textbf{74.47}(+2.76)   &  0.54   \\
\cline{2-8}
                          & \multirow{3}*{50\% / 60\% / 70\%}& Scratch    & 72.76   & 70.60                   & - & - &-  & -      \\
                          &                       & Label      & 74.82    & 72.43                   & 71.71        & 73.71  & \textbf{73.97}(+3.37)  & 1.31 \\
                          &                       & Feature    & 74.98    & 71.98                   & 71.26        & 73.42  & \textbf{73.71}(+3.11)  & 0.89   \\

\Xhline{3\arrayrulewidth}
\end{tabular}
}
\end{center}
\caption{Test accuracy (\%) of classification networks that used $L_{1}$-norm based channel pruning on CIFAR-100 dataset. The `Train Type' column indicates whether the network is trained using the label distillation or feature distillation or without distillation (scratch). The `Remaining Channels' column {indicates} the percentage of remaining channels at each ResNet groups.}
\label{table:1}
\end{table*}

\section{{KD} with Pruning}
\label{sec:4}

To examine the orthogonal property of KD with other efficiency-improving methods, we first choose network pruning as it is one of the most representative compression methods. Given that a model trained with KD functions as network ensemble, we expect that KD will always be beneficial to the pruned network.
{Note that} there are already some studies that propose to use KD with network pruning \cite{bao2019using,hu2018novel}. 
However, earlier works only combine KD at the fine-tuning process {where a very small learning rate is used}.
We argue that distilling only at the fine-tuning step is overlooking an important role KD plays at the initial training phase leading to only a marginal improvement, as will be presented in our results.

To analyze the interactions of the two methods, we apply KD to various network pruning methods at different phases on CIFAR-100 dataset. In this experiment, KD can be applied at two phases: 1) at the initial training phase of the network before pruning  (\textit{Pre-Distill}\footnote{Pre-Distill results are presented in the same column with Finetuned for scratchd networks for all tables.}), 2) after pruning at the fine-tuning phase (\textit{Post-Distill}). {We also apply distillation at both phases referred to as \textit{Pre-Post}.} {Additionally,} at Pre-Post, instead of using a teacher network at Post phase, {an unpruned student network can be} used as the teacher network. We name this {as} \textit {Self-Distill}. 

\subsection{Structured Pruning}
\textbf{$L_{1}$-norm based Filter Pruning \cite{li2016pruning}} is an early work on structured pruning. For each layer, {a} heuristically pre-defined percentage of filters {having} a small $L_{1}$-norm are pruned. The {pruning} ratio can be different for each layer. ResNet \cite{he2016deep} for example, has different pruning ratios for each group, and prevents certain convolution layer from being pruned in order to prevent dramatic accuracy drop. 
The results are reported on Table \ref{table:1}. 

\noindent \textbf{Network Slimming \cite{liu2017learning}} imposes $L_{1}$-norm penalty on convolution layers during the initial training {phase} of the network, so that the network can be pruning-friendly. 
There are several studies that try to sparsify a network at weight-level. 
However, Network Slimming imposes sparsity on channels and achieves higher performance compared to previous works.
The results of pruning combined with KD are reported in Table \ref{table:2}.

\begin{table*}[ht]
\begin{center}
\resizebox{1.0\linewidth}{!}{
\begin{tabular}{c|c|c|c|cccc|c}
\Xhline{3\arrayrulewidth}
Model Type                & Remaining channels  & Train Type   & Unpruned         & Finetuned / Pre-Distill  & Post-Distill & Pre-Post       & Self-Distill           & Surplus \\
\cline{1-9}
 \multirow{3}*{VGG-19}    & \multirow{3}*{50.0\%} & Scratch    & 72.03            &  72.30                   &   -          &   -            &    \\
                          &                       & Label      & 73.84(+1.81)     &  73.63                   & 72.36        & 73.80          & \textbf{73.91}(+1.61)  & -0.20  \\
                          &                       & Feature    & 73.89(+1.86)     &  73.06                   & 72.41        & 73.91          & \textbf{74.21}(+1.91)  &  0.05  \\
\cline{1-8}
 \multirow{3}*{ResNet-56}& \multirow{3}*{50.0\%}  & Scratch    & 73.07            &  72.45                   &   -          &   -            &      \\
                          &                       & Label      & 74.52(+1.45)     &  74.32                   & 73.25        & 74.39          & \textbf{74.78}(+2.33)  & 0.88   \\
                          &                       & Feature    & 74.63(+1.56)     &  73.89                   & 73.62        & 74.37          & \textbf{74.54}(+2.09)  & 0.53   \\
\cline{1-8} 
 \multirow{3}*{DenseNet-40}& \multirow{3}*{30.0\%}& Scratch    & 73.14            &  72.30                    & -           & -              & -   & \\   
                          &                       & Label      & 74.96(+1.82)     &  74.32                    & 73.64       & \textbf{74.59} & 74.31(+2.01)           & 0.19   \\
                          &                       & Feature    & 74.19(+1.05)     &  74.17                    & 73.81       & 74.40          & \textbf{74.43}(+2.13)  & 1.08   \\

\Xhline{3\arrayrulewidth}
\end{tabular}
}
\end{center}
\caption{Test accuracy (\%) of classification networks that used Network Slimming on CIFAR-100 dataset.}
\label{table:2}
\end{table*}

\begin{table*}[ht]
\begin{center}
\resizebox{1.0\linewidth}{!}{
\begin{tabular}{c|c|c|c|cccc| c}
\Xhline{3\arrayrulewidth}
Model Type                & Remaining Params      & Train Type & Unpruned    & Finetuned/Pre-distill & Post-Distill & Pre-Post          & Self-Distill                 & Surplus \\
\cline{1-9}
 \multirow{3}*{ResNet-110}& \multirow{3}*{40.0\%} & Scratch    & 72.82       & 72.02                 & -            & -                 &    -                         & - \\
                          &                       & Label      & 75.01(+2.19)& 73.38                 & 73.27        & \textbf{75.08}    & 74.85(+2.83)                 & 0.64\\
                          &                       & Feature    & 74.41(+1.59)& 73.08                 & 73.29        & 74.02             & \textbf{74.20}(+2.18)        & 0.59 \\

\Xhline{3\arrayrulewidth}
\end{tabular}
}
\end{center}
\caption{Test accuracy (\%) of networks that used unstructured weight-level pruning with KD on CIFAR-100 dataset.}
\label{table:3}
\end{table*}
\subsection{Unstructured Pruning}
\textbf{Weight-level Pruning \cite{han2015deep}} performs unstructured magnitude-based weight pruning on individual weights. 
We only apply pruning to the convolution layers, meaning that the final layer (which maps pooled feature map logits using single linear operation) is not pruned. The scores are reported on Table \ref{table:3}.

For all three methods, a number of similar tendencies could be found. 
First, as expected by the orthogonal property, applying KD leads to a noticeable performance boost at all stages. Applying KD at the initial training phase (\textit{Pre-Distill}) usually outperformed \textit{Post-Distill}, and \textit{Pre-Post} always outperformed the two schemes. 
Moreover, \textit{Self-Distill} performed the best in most settings. 
{This {hints at a practical scheme}} of combining KD with pruning {for} stronger performance.

In the last column of the three tables, we calculated the surplus gain to quantify the extra performance gain yielded by KD. 
This is calculated by using the Self-Distill, Finetuned, and Unpruned columns. 
In detail, surplus is defined as the difference between the performance gain of KD when applied to a \textit{scratch network} and performance gain of KD \textit{when combined with the pruning and finetuning process.} 
The former can be computed by (Unpruned KD - Unpruned Scratch), while the latter is (Finetuned KD - Finetuned Scratch).  
For convenience, we only used the results of Self-Distill for Finetuned KD.
If the performance gain of KD does not conflict with the performance recovery of fine-tuning the pruned network, we expect that the surplus is 0.
For detailed example, in Table \ref{table:1}, ResNet-56 with top three rows, the surplus on label distillation is calculated by [(Self-distill - finetuned) - (Label Unpruned - Scratch Unpruned)] = [(73.30 - 70.06) - (73.76 - 71.23)] = 0.71.
In nearly all cases, KD yielded positive surplus, while some even surpassing 1\%, showing not only the independence but also a clear synergy of KD and pruning.


\section{{KD} with Image-level Regularization}
\label{sec:5}
Let us denote the data augmented by Image-level regularization as $D_{I}$.
As noted on the {Section \ref{sec:kd}}, we claimed that KD boosts the effective number of training samples, and thus, resembles the properties of network ensemble or bagging.
Therefore, it is questionable whether their role may overlap with {data} augmentation methods. 
We claim that this is {partly} true, but we can still expect decent accuracy gain for following reason{s}:
{1)} KD acts as a data augmentation on the output level with fixed input images, whereas image-level regularization operate{s} on the input level where the inputs change, but the output labels are usually fixed. Thus, their augmentation effects act on different level{s}.
2) Since $D_{A}$ is defined to include all the possible augmented samples, {$D_I$} is also a subset of $D_A$. Thus,
we can expect them to be orthogonal to the augmentation effect of KD.

In this section, 
we {demonstrate the results of combining} KD with two different kind{s} of image-level regularization, \textit{Self-Augment} and \textit{Cross-Augment}.
One can decide whether to apply regularization to the student and/or teacher, with four possible combinations in total: 1) Both the teacher and student are trained without regularization, which is equivalent to VKD, 2) The teacher is trained with normal samples, but the student is trained with regularization. In this case, the teacher is not exposed to the regularization, but regularization-applied samples are forwarded into both the teacher and the student to extract class probability or feature maps. 3) The teacher is trained using regularization-applied samples, but the vanilla samples are forwarded to both the teacher and student. 4) Both student and teacher are trained and forwarded with regularization-applied samples.

\begin{table}[ht]
\begin{center}
\resizebox{1.0\linewidth}{!}{
\begin{tabular}{c|cc|cc}
\Xhline{3\arrayrulewidth}
\multirow{2}*{CIFAR-100}& \multicolumn{2}{c|}{AutoAug Trained} & \multirow{2}*{Label}& \multirow{2}*{Feature} \\
\cline{2-3}
& Teacher & Student &                          &                \\
\cline{1-5}
\multirow{2}*{Pyramid-200-300}                & No      & No               & 83.96          & 84.45\\
                                              & No      & Yes              & 86.84          & 86.18\\
\cline{1-1}
{Scratch: 83.46}                              &Yes      & No               & 85.32          & 86.49\\      
{AutoAugment: 85.22}                          &Yes      & Yes              & 87.27          & 86.61\\ 
\Xhline{3\arrayrulewidth}
\end{tabular}
}
\end{center}
\caption{Test classification accuracy (\%) of AutoAugment with KD on CIFAR-100.}
\label{table:4}
\end{table}

\subsection{Self-Augment}
\textbf{AutoAugment \cite{cubuk2018autoaugment}} proposes learning the optimal data augmentation policies with reinforcement learning. 
They achieved state-of-the-art accuracy on CIFAR-10, CIFAR-100 \cite{krizhevsky2009learning}, SVHN \cite{37648}, and ImageNet \cite{ILSVRC15} by applying Shake-Shake regularization \cite{gastaldi2017shake} or Shake-Drop \cite{yamada2019shakedrop} regularization together. 
For the augmentation policy, we chose the random policy of AutoAugment whose CIFAR-10 classification error is 3.0\% in WideResNet-28-10 \cite{zagoruyko2016wide} which is only 0.4\% worse than the best policy (2.6\%).
The result of PyramidNet \cite{han2017deep} on the CIFAR-100 dataset is on Table \ref{table:4}.

The relative accuracy gain compared to scratch training of random policy AutoAugment is 1.76\%, and the gain of label distillation and feature distillation is 0.50\% and 0.99\% respectively.
The accuracy gain of random AutoAugment + label distillation and random AutoAugment + feature distillation is 3.81\% and 3.15\%, which is much higher than the sum accuracy gain of each cases, 2.24\% and 2.73\%.
These imply that KD is not limited to boosting performance independently with AutoAugment, but even has a synergistic interaction.  
These phenomena are impressive, given that performance gain usually diminishes as the baseline performance gets higher.
When combined with AutoAugment at both stages of teacher and student training (yes/yes), 
the label distillation performs better than the feature distillation.

\subsection{Cross-Augment}
\textbf{CutMix Regularization \cite{yun2019cutmix}} proposes to mix the training samples in a patch-wise manner by randomly cropping and swapping them. 
The detailed explanations can be found in Eq. (\ref{eq2}).
As shown by the equations, they not only mix the training images but also mix the labels proportionally.
We combine KD and CutMix in the same manner as in AutoAugment. 
The results are reported on Table \ref{table:5}.

\begin{table}[t]
\begin{center}
\resizebox{1.0\linewidth}{!}{
\begin{tabular}{c|cc|cc}
\Xhline{3\arrayrulewidth}
\multirow{2}*{CIFAR-100}& \multicolumn{2}{c|}{Cutmix Trained}  & \multirow{2}*{Label}     & \multirow{2}*{Feature} \\
\cline{2-3}
                        & Teacher & Student                    &                          &                        \\
\cline{1-5}
\multirow{2}*{Pyramid-110-64}                  & No      & No         & 81.92               & 81.25\\
                                               & No      & Yes        & 82.31               & 83.19\\
\cline{1-1}
{Scratch: 80.27 / 80.15*}                      &Yes      & No         & 82.40               & 83.13\\      
{Cutmix: 82.09 / 82.03*}                       &Yes      & Yes        & 83.72               & 83.35\\ 
\cline{1-5}
\multirow{2}*{Pyramid-200-240}                 & No      & No         & 84.34               & 84.56\\
                                               & No      & Yes        & 85.43               & 86.08\\
\cline{1-1}
{Scratch: 83.41 / 83.55*}                      &Yes      & No         & 83.52               & 85.53\\      
{Cutmix: 84.87 / 85.53*}                       &Yes      & Yes        & 86.39               & 86.31\\ 
\Xhline{3\arrayrulewidth}
\end{tabular}
}
\end{center}
\caption{Test classification error (\%) of Cutmix with KD on CIFAR-100. The numbers with * on the leftmost column are the network's reported score on the corresponding paper.}
\label{table:5}
\end{table}

For PyramidNet-200-240, we were unable to reproduce the reported result using the official code, achieving 0.67\% worse accuracy. 
Both our result and the official result are reported.
On PyramidNet-110-64, the relative accuracy gains of CutMix, label distillation, and feature distillation are 1.82\%, 1.65\%, and 0.98\% each, but when combined, the accuracy gain of label distillation + CutMix and feature distillation + CutMix are 3.45\% and 3.08\%, which are similar to the sums of each gain, 3.47\% and 2.80\%.
On PyramidNet-200-240, the accuracy gain of CutMix, label distillation, and feature distillation is 1.46\%, 0.93\%, and 1.15\% each, but when combined, the accuracy gain of label distillation + CutMix and feature distillation + CutMix are 2.98\% and 2.90\%, which are better than the sums of each gain, 2.39\% and 2.61\%.
The results of KD + CutMix (86.39\%) is better than CutMix + ShakeDrop, which was reported to be 86.19\% on CIFAR-100.

One more point worth noting is that when the trained environment and distilled environment of the teacher network and the student network are different, as in No / Yes row and Yes / No row, we observed that the test accuracy becomes unstable.
Interestingly, we found that for CutMix, when the trained environment and deployed environment are different (No / Yes row and Yes / No row), label distillation suffers. 
We infer that these results stem from the distributional difference. 
Intuitively, the distributional discrepancy between the original dataset and the augmented dataset is larger for Cross-Augment than is for Self-Augment as both the input and target distributions change for the former.

\section{Qualitative analysis}
\label{sec:6}
\subsection{Measuring the Representation Diversity}
We analyzed KD in the perspective of augmenting the effective number of samples, and concluded that KD is analogous to ensemble, especially bagging. As model ensemble benefits from having diversified outputs of separate models, a model that has successfully attained knowledge from multiple networks will have relatively diverse outputs. 
Thus, we expect that given a same batch of inputs, a model trained with KD will return more diversified outputs than a model trained from scratch.
To assess the diversity patterns of outputs, we adopted a recent metric from NAS Without Training \cite{mellor2020neural}.
The proposed metric is computed from the correlation matrix of the approximated linear operators \cite{hanin2019deep} that map each data point to the output. 
This metric assigns higher scores to models with lower correlation given a mini-batch.
Having low correlation between output patterns suggests that the model's output patterns are diverse, and we expected KD to result in higher diversity in output patterns.
We averaged the scores of five independently trained models on CIFAR-10, and the results are in Table \ref{table:6} (the higher the number, the more diverse the output).
With the 99\% confidence interval, we can conclude that the KD scores significantly higher than the scratch model, and feature distillation + CutMix scores little higher, meaning that these methods do contribute to making a fixed sized network output diverse representations. 
More studies about the output patterns of a neural network can be found in studies such as \cite{zhang2016understanding,arpit2017closer}.

\begin{table}[t]
\begin{center}
\resizebox{1.0\linewidth}{!}{
\begin{tabular}{c|ccccc}
\Xhline{3\arrayrulewidth}
ResNet-56            & Scratch & CutMix & Label   & Feature  & Feat+CutMix \\
\Xhline{1\arrayrulewidth}
Avg                  & -143.96 & -143.11 & -142.22 & -141.94 & -141.34 \\
StdDev               & 0.462   & 0.616   & 0.560   & 0.353   & 0.186 \\
ErrMargin            & 0.207   & 0.275   & 0.250   & 0.157   & 0.092 \\
99\% interval        & 0.532   & 0.709   & 0.645   & 0.406   & 0.239 \\
 \Xhline{3\arrayrulewidth}
\end{tabular}
}
\end{center}
\caption{Measured scores for different methods. At fourth column for example, the average score of label distillation is -142.22 with the 99\% confidence interval being $\pm0.645$}
\label{table:6}
\end{table}

\begin{table*} [ht]

  \begin{tabular}{|c|c|c|c|} 
  \Xhline{3\arrayrulewidth}
  CIFAR-10 & First Decay Step & Second Decay Step & Third Decay Step\\
      \Xhline{2\arrayrulewidth}
      {Scratch} &
      \parbox[c]{13.6em}{\includegraphics[width=1.2in]{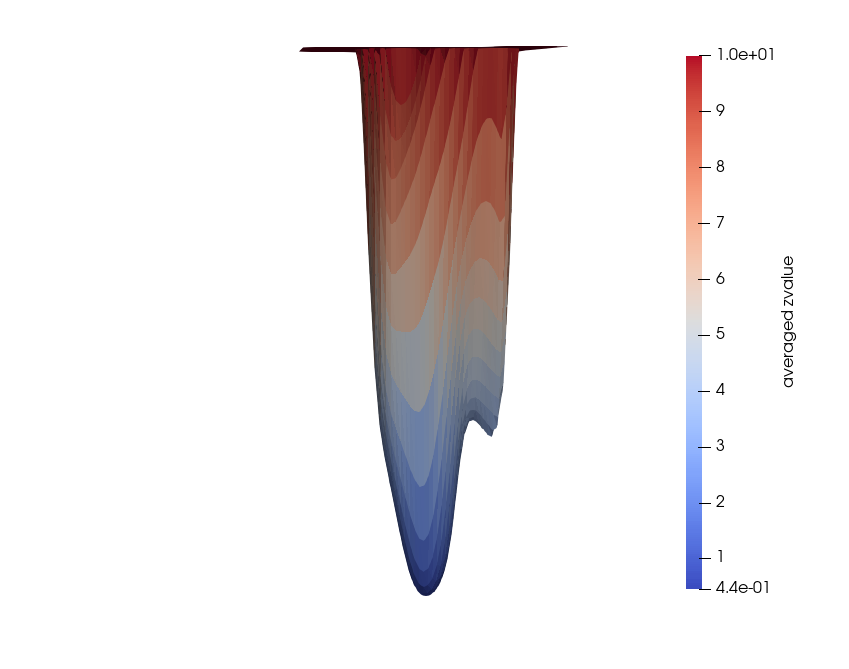}} & 
      \parbox[c]{13.6em}{\includegraphics[width=1.2in]{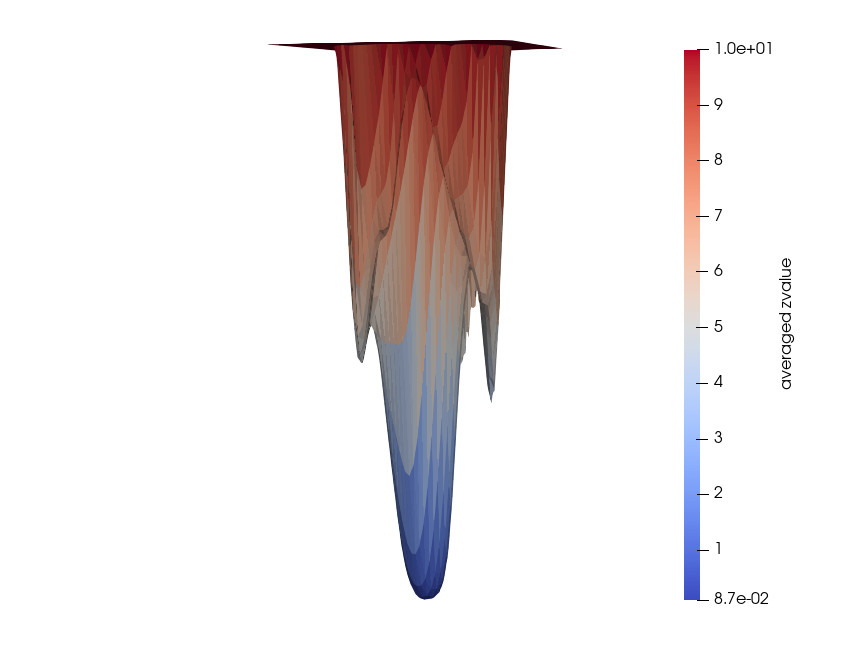}} & 
      \parbox[c]{13.6em}{\includegraphics[width=1.2in]{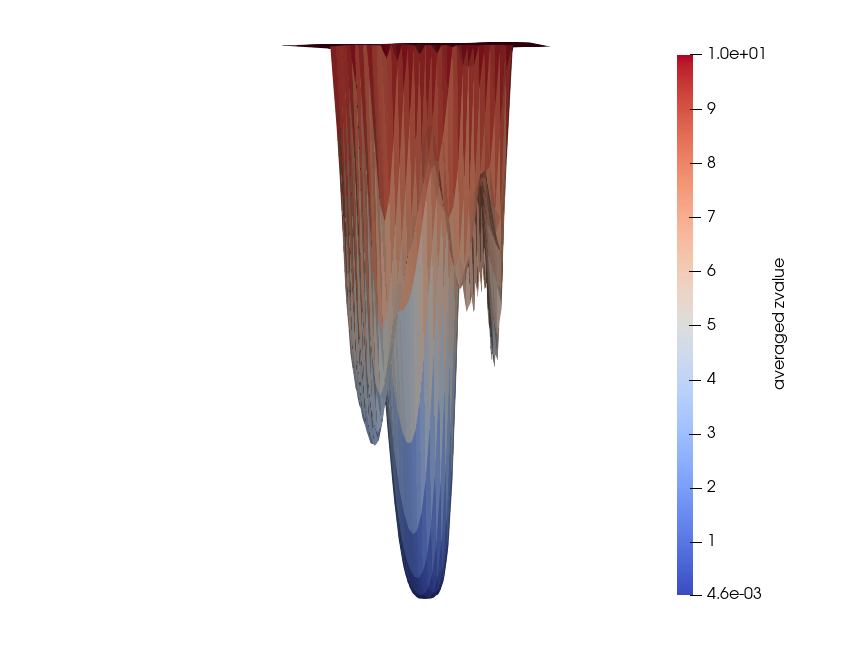}}\\
      \Xhline{2\arrayrulewidth}
      {CutMix} &
      \parbox[c]{13.6em}{\includegraphics[width=1.2in]{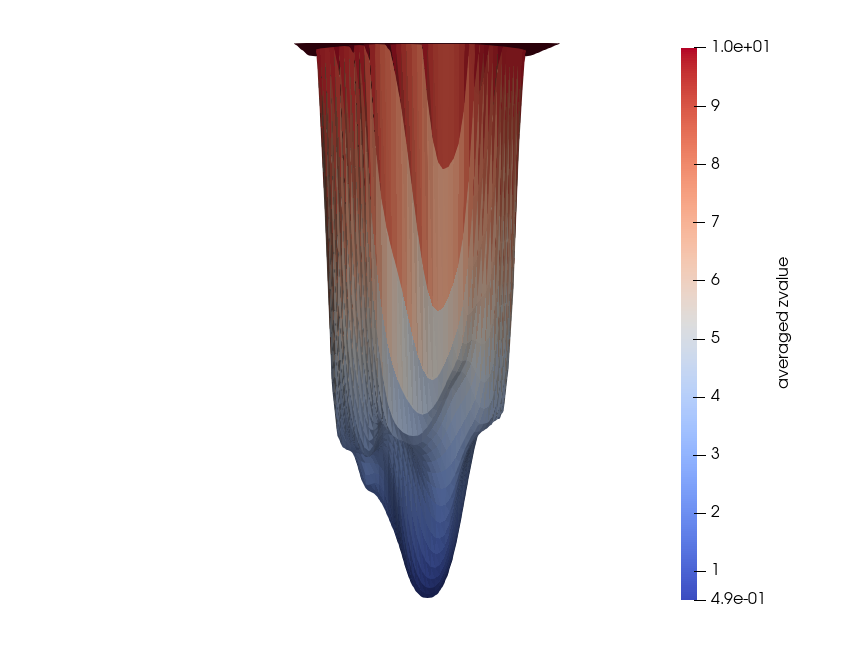}} & 
      \parbox[c]{13.6em}{\includegraphics[width=1.2in]{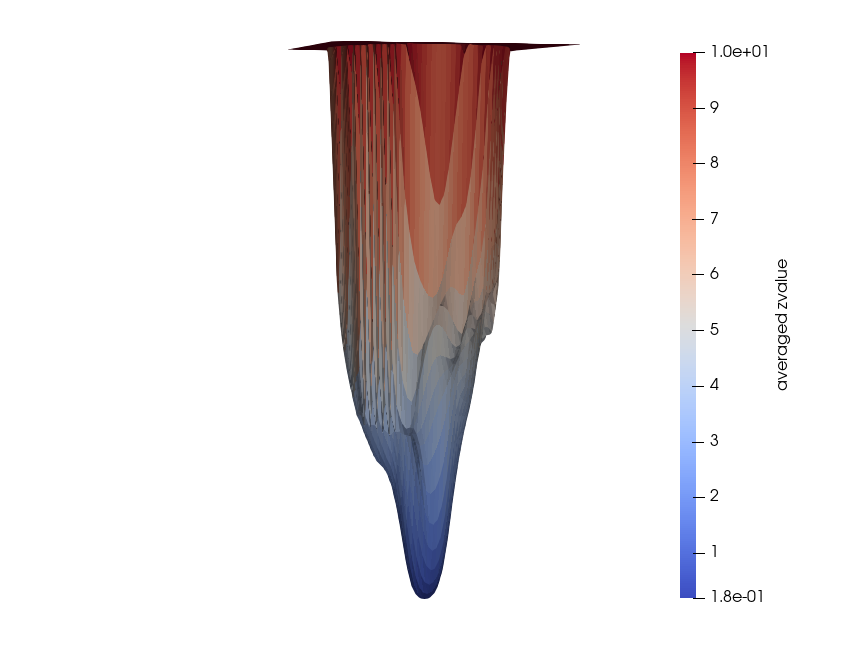}} & 
      \parbox[c]{13.6em}{\includegraphics[width=1.2in]{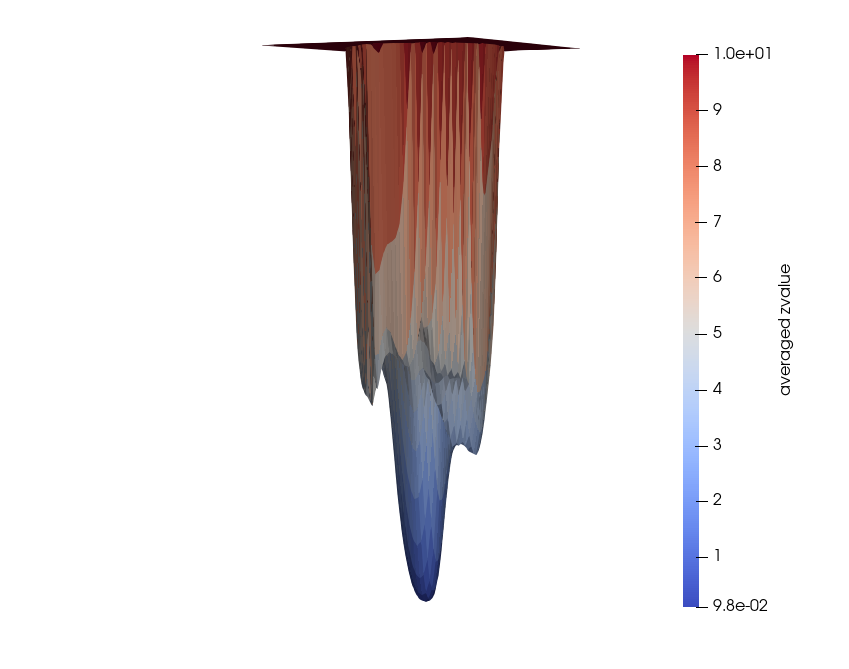}}\\
      \Xhline{2\arrayrulewidth}
      {Distill} & 
      \parbox[c]{13.6em}{\includegraphics[width=1.2in]{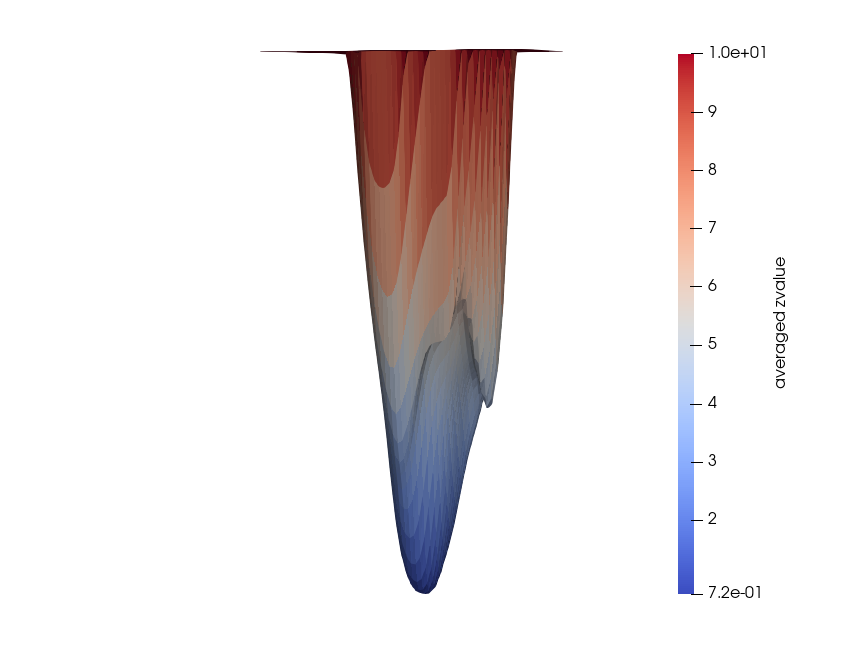}} & 
      \parbox[c]{13.6em}{\includegraphics[width=1.2in]{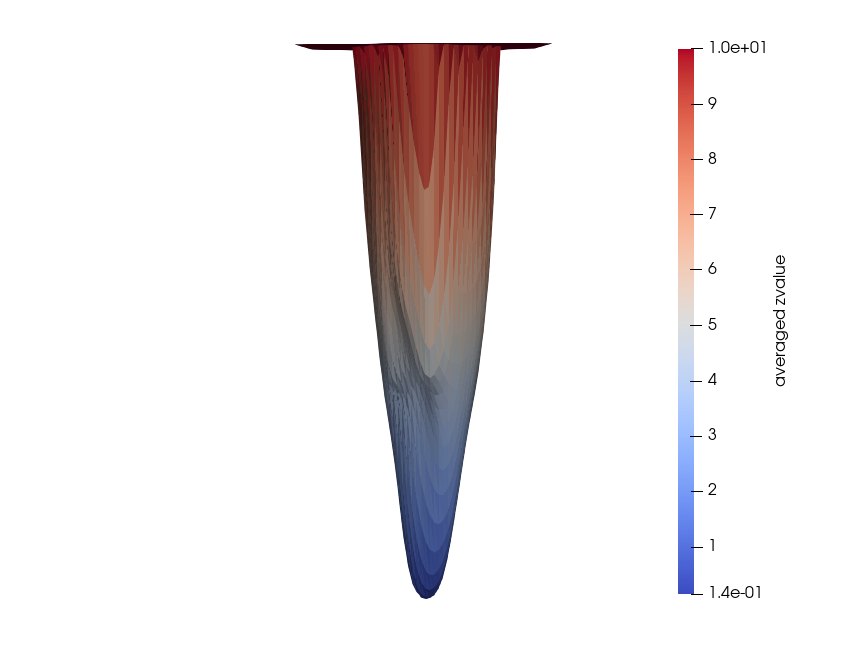}} & 
      \parbox[c]{13.6em}{\includegraphics[width=1.2in]{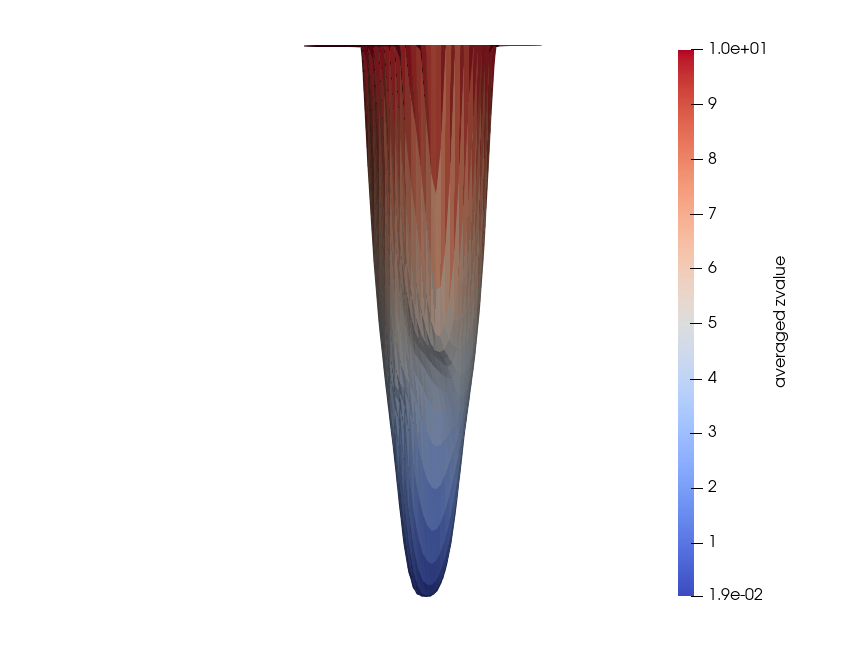}}\\
      \Xhline{3\arrayrulewidth}
  \end{tabular}
 \caption{Loss landscapes of the ResNet-56 for Scratch trained model, CutMix model, and Distillation model. In order to take a closer look at sharpness of the local minima, an enlarged view using PDF version is recommended. }
 \label{losssurface}
\end{table*}

\subsection{Visualizing the effect of knowledge distillation}
To qualitatively assess how KD behaves differently from other efficiency-enhancing methods, we visualized and compared the loss landscape of the training process in Table \ref{losssurface}. Additionally, the video rendering the landscape is included in the supplementary material.
At the end of each three learning rate step, we plotted the loss surfaces for three training processes: scratch training, CutMix training, and KD training.
When the three landscapes are compared to each other, several aspects could be observed starting from the early learning step: 
\begin{itemize}
    \item 
    The scratch model training makes many different sharp modes (minima) which consequently makes sharp optima. This may be due to overfitting to the training dataset. 
    \item The optimization of CutMix regularization takes place in a much wider landscape. This may be because CutMix allows the loss function to explore more diverse zones -- namely, the convex combination of the training samples.
    \item KD yields a visibly smoother minimum from the first decay step, and as training progresses it draws the different minima together, which eventually converge into a single smooth one. 
    This is in stark contrast with the other methods that have multiple local minima surrounding one main minimum.
\end{itemize}

We add more discussions about these observations on the next section.

\section{Discussion}
\label{sec:7}
Intuitively, the smoothing effect of KD can be explained by the effect of model ensemble which learns averaged outputs as a target, making the predicted class probabilities smoother, leading to robustness to noises.
This hints at the effectiveness of KD in defense against adversarial attacks which have been empirically reported in \cite{papernot2016distillation}, while \cite{Zhao2020Bridging} independently reports that smooth loss landscape leads to robustness against adversarial attacks. 
Our experimental results may help further interpretation of the results in \cite{papernot2016distillation}. 
Also, different aspects appearing at the early learning step have connection with ESKD \cite{Cho_2019_ICCV}. 
Additionally, DML \cite{zhang2017deep} states that KD makes wider minima that generalize better, which supports our third observation. 

Next, we discuss two more topics. 
While the results of our experiments support the orthogonality of KD with some efficiency-improving methods, other works exist that have not been demonstrated experimentally. Therefore, we first introduce past findings from other works that empirically support the orthogonal property of KD.
For instance, the coefficient of weight decay regularization ($L_2$-penalty) is a hyperparameter that depends on the baseline methods. %
The accuracy of scratch networks is reported to be highly affected by the weight decay value \cite{he2016deep,zagoruyko2016wide,xie2017aggregated}. 
But the reported scores on various papers show that KD consistently boosts the performance no matter which value is used.
In addition, `compressed blocks', architectures that enhances efficiency, have also been covered by past works.
Moonshine distillation \cite{crowley2018moonshine} reports that VKD and AT still enhances the performance of networks composed of `compressed blocks' such as group convolution or bottleneck layer, supporting the orthogonal property of KD. 

Second, we explain a well-reported phenomenon that is difficult to interpret with the standard viewpoint of KD. Many papers have reported that the performance increases even when the student network is of the same kind or larger than the teacher network. Moreover, in some cases a small-capacity student network exceeds the teacher network. This contradicts the intuition of KD that teacher network injects knowledge to an inferior student network.
Such results have often been used to support the superiority of the methodologies \cite{furlanello2018born,chung2020feature,zhang2017deep}. 
However, these results are natural from the viewpoint of KD functioning as model ensemble. 
The teacher network is not simply teaching the student network: 
the student network is trained to mimic the ensemble. 

\section{Conclusion}
\label{sec:conclusion}
In this paper, we explained the behavior of KD as ensemble (bagging) by reinterpreting it from the sample augmentation perspective.
As model ensemble generally enhances performance regardless of model architecture or the underlying method, we expected KD to show a similar tendency. We applied KD for network pruning and regularization methods, and
experimentally demonstrated the versatility of knowledge distillation as a generally applicable approach to construct an efficient model with promising performance boost. 
In both cases, it is worth noting that we only used primitive forms of label distillation and feature distillation shown in Eq. (\ref{eq:LKD}) and (\ref{eq:LFD}). 
Better results may be achievable with rather recent and superior distillation methods. 
We hope our findings to be spread and discussed broadly, so that many researchers and practitioners find knowledge distillation as worthy of being studied and utilized eagerly.

\appendix
\section{Experiment Settings For reproduction}
\subsection{Baseline Codes}
The baseline codes for feature distillation and label distillation for 
CIFAR-100 are {in} the code appendix. Since the purpose of our experiments is to empirically verify the orthogonal performance enhancement of KD with ``Other Techniques", we conducted all the experiments on the {publicly released} github code of other papers. 

For the network pruning, we applied KD on {the methods implemented on} \textit{https://github.com/Eric-mingjie/rethinking-network-pruning/tree/master/cifar}. They support 5 branches of pruning methods. For channel pruning we used ``l1-norm-pruning"  and ``network slimming" branches, and for unstructured pruning, we used ``weight-level pruning" branch.

For image-level regularization, we experimented {with} Self-Augment and Cross-Augment.
For Self-Augment, we used {the} random policy of AutoAugment \cite{cubuk2018autoaugment} {whose} code is on \textit{https://github.com/DeepVoltaire/AutoAugment}. The usage is on ``README.md", and Class module for random policies are in ``autoaugment.py".
For Cross-Augment, CIFAR-100 setting of CutMix \cite{yun2019cutmix} {whose} code is on \textit{https://github.com/clovaai/CutMix-PyTorch} (Copyright (c) 2019-present NAVER Corp.
). We {used} the minibatch-level image manipulating method of CutMix and {the} dataloader of AutoAugment on other codes for experiments.

On the Qualitative Analysis section, {we first }showed the score of the trained networks using the metric of NAS Without Training \cite{mellor2020neural}. The official code is on \textit{https://github.com/BayesWatch/nas-without-training}. {W}e only adopted the \textit{get\_batch\_jacobian} function and \textit{eval\_score} function on ``search.py". The two function were easy to plug-in {to other methods.}

Second, we used {the} Paraview tool to visualize the loss surface. The codes for {outputting} the surface files and the link for the Paraview program are on \textit{https://github.com/tomgoldstein/loss-landscape}. The usages are kindly explained on the ``README.md".

\subsection{Hyperparameters for Distillations}
On VKD \cite{hinton2015distilling}, {w}e used the balancing parameter $\alpha$ of 0.9 and the temperature $T$ of 4. For feature distillation, the choice of $\beta$ is {not trivial}. It has {a} positive correlation with the number of elements ($H\times W\times C$), and we use it to approximately balance the loss value between the CE loss and distillation loss at the beginning step of the training. 
For our experiments on network pruning, we used $\beta$ of 500 for ResNets \cite{he2016deep}, and 1500 for VGGs \cite{DBLP:journals/corr/SimonyanZ14a} and 1000 for DenseNet-40 \cite{huang2017densely}. 
On Cross-Augment and Self-Augment, we used $\beta$ of 1000 for PyramidNet-110-64 \cite{han2017deep} and 2000 for PyramidNet-200. {For a different kind of network, $\beta$ needs to be re-scaled to an appropriate value.}

\begin{table*} [h]
  \begin{tabular}{|c|c|c|c|} 
  \Xhline{3\arrayrulewidth}
  CIFAR-10 & First Decay Step & Second Decay Step & Third Decay Step\\
      \Xhline{2\arrayrulewidth}
      {Scratch} &
      \parbox[c]{12em}{\includegraphics[width=1.8in]{scr70.png}} & 
      \parbox[c]{12em}{\includegraphics[width=1.8in]{scr110.png}} & 
      \parbox[c]{12em}{\includegraphics[width=1.8in]{scr160.png}}\\
      \Xhline{2\arrayrulewidth}
      {CutMix} &
      \parbox[c]{12em}{\includegraphics[width=1.8in]{cutmix70.png}} & 
      \parbox[c]{12em}{\includegraphics[width=1.8in]{cutmix110.png}} & 
      \parbox[c]{12em}{\includegraphics[width=1.8in]{cutmix160.png}}\\
      \Xhline{2\arrayrulewidth}
      {Label Distill} & 
      \parbox[c]{12em}{\includegraphics[width=1.8in]{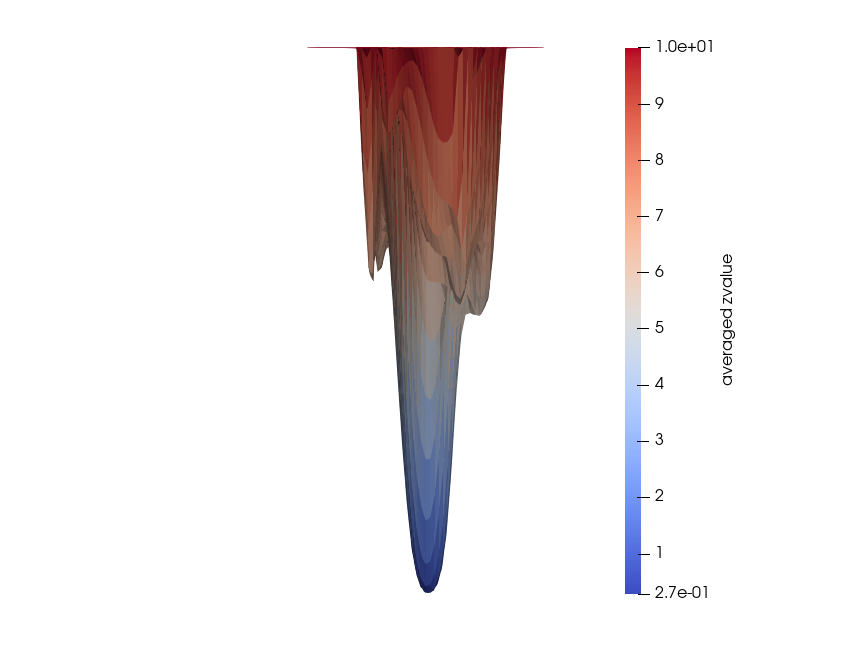}} & 
      \parbox[c]{12em}{\includegraphics[width=1.8in]{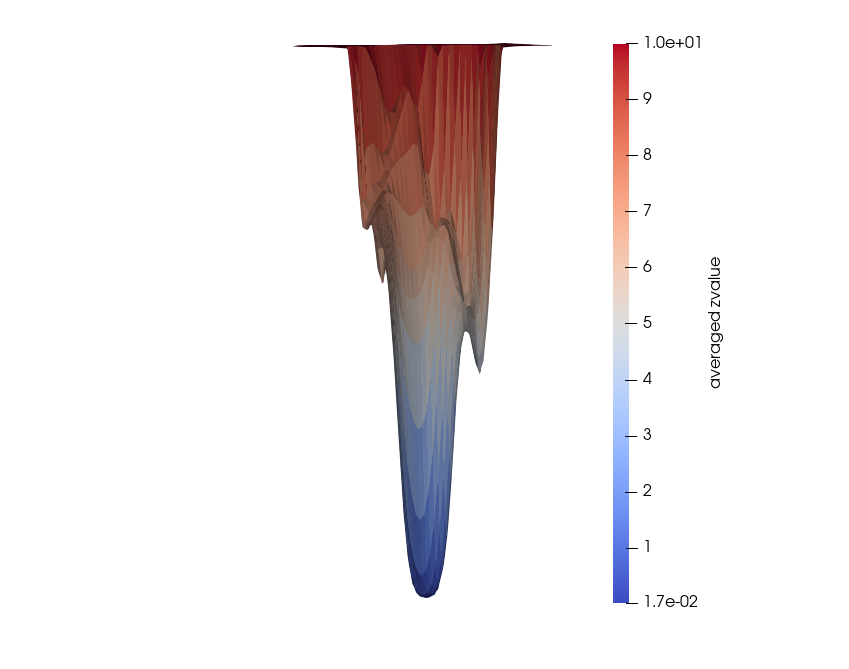}} & 
      \parbox[c]{12em}{\includegraphics[width=1.8in]{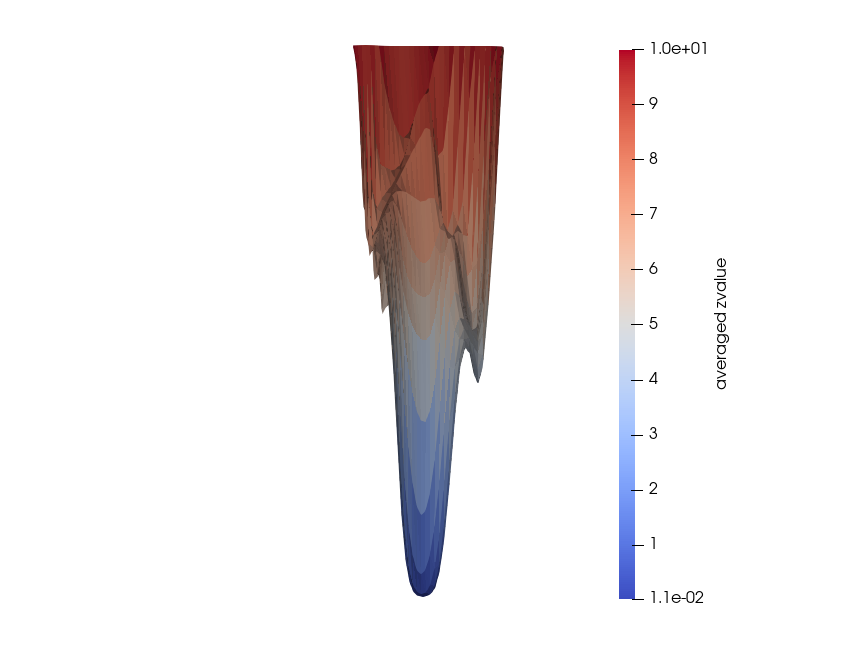}}\\
      \Xhline{2\arrayrulewidth}
      {Feature Distill} & 
      \parbox[c]{12em}{\includegraphics[width=1.8in]{distill70.png}} & 
      \parbox[c]{12em}{\includegraphics[width=1.8in]{distill110.png}} & 
      \parbox[c]{12em}{\includegraphics[width=1.8in]{distill160.png}}\\
      \Xhline{2\arrayrulewidth}
      {Cutmix+Feature Distill} &
      \parbox[c]{12em}{\includegraphics[width=1.8in]{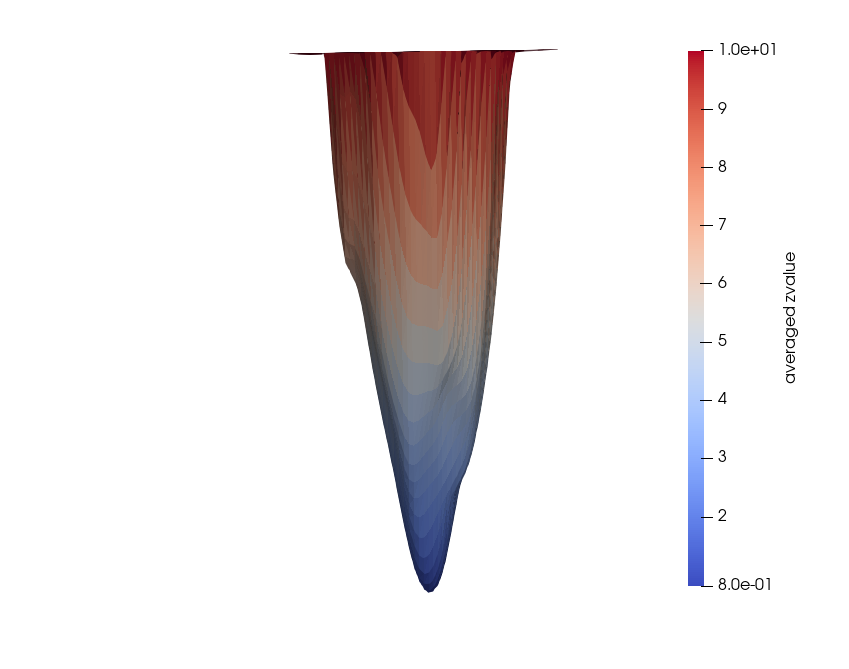}} & 
      \parbox[c]{12em}{\includegraphics[width=1.8in]{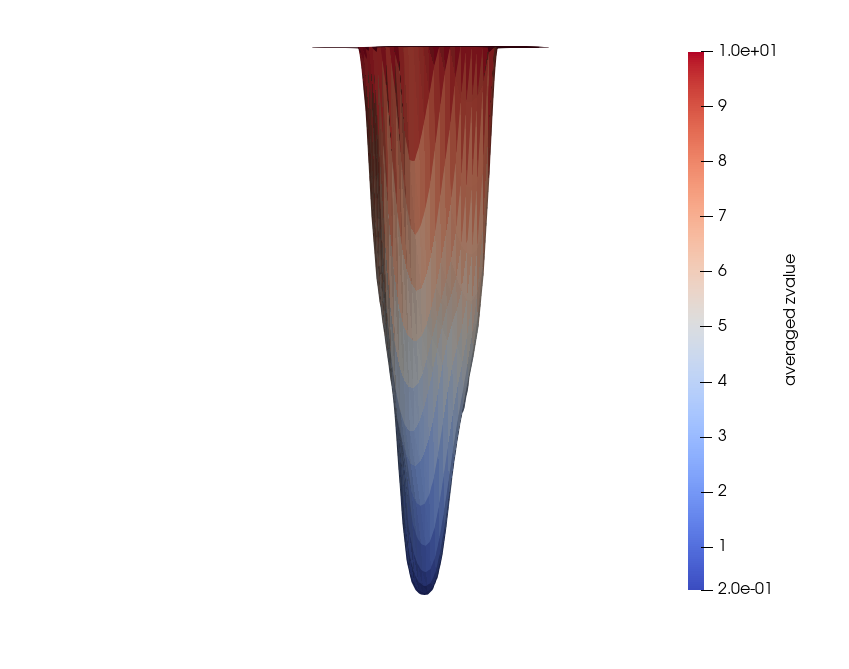}} & 
      \parbox[c]{12em}{\includegraphics[width=1.8in]{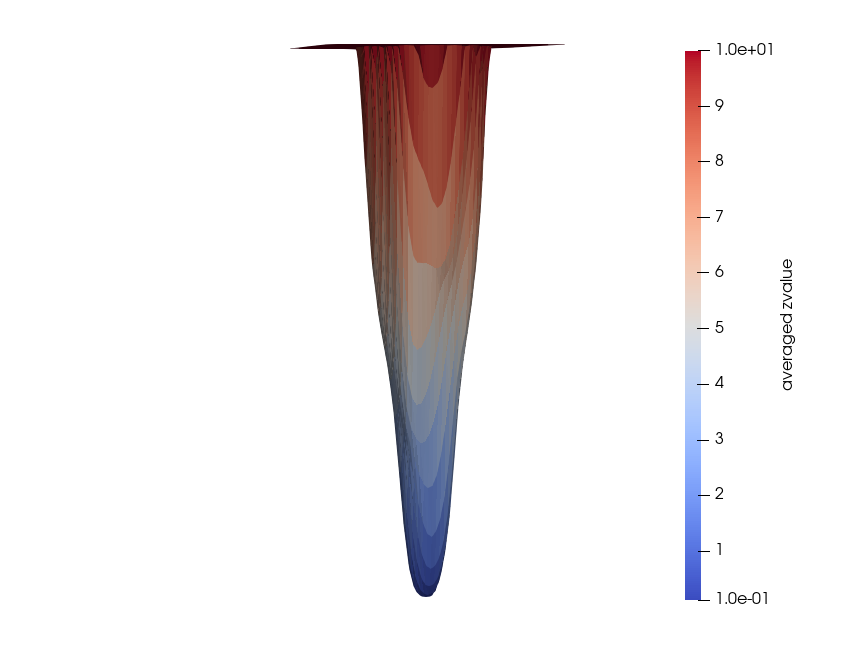}}\\
      \Xhline{3\arrayrulewidth}
  \end{tabular}
 \caption{Loss landscapes of the ResNet-56 for various training methods. In order to take a closer look at sharpness of the local minima, an enlarged view using PDF version is recommended. }
 \label{surface}
\end{table*}

\section{Interpretation of Feature Distillation from an Ensemble Perspective}
As we explained on Section 3 in our paper, for VKD the softmax output of the teacher network $D_T$ can be interpreted as an augmented dataset at label-level. {It can be} further interpreted as $D_{T}'$ {with an} increased effective number of samples. 
For feature distillation, there only exist linear operations between the feature map activation and the softmax {outputs and the student is induced to reproduce the activation map of the teacher.}
Through the linear operation, each class takes the learned weighted sum of global-averaged channel activation, and the weighted sum of the last channel activation produced by the linear operation become the logits for each classes.
Thus, each channel may contain fine{r} information that are less refined (less abstract information).
{U}sing loss for features may have {an} effect of augmenting by giving more specific targets to learn. The target of the label is N-hot vector, whereas the target of feature is {H}$\times${W}$\times${C} vector.

\section{More Visualization of loss landscape}
We {additionally} plotted loss landscapes \cite{li2018visualizing} at Table \ref{surface}. Interestingly, the feature distillation definitely looks much smoother than VKD. We infer the reason for this phenomenon is that feature distillation accompany more detailed targets. We also included a video file {visualizing the landscapes from various perspectives for a detailed comparison} in our supplementary materials.

\clearpage
\bibliography{2021Main}
\end{document}